\documentclass[conference]{IEEEtran}
\IEEEoverridecommandlockouts
\usepackage{booktabs}
\usepackage{cite}
\usepackage{amsmath,amssymb,amsfonts}
\usepackage{algorithmic}
\usepackage{graphicx}
\usepackage{textcomp}
\usepackage{xcolor}
\usepackage{orcidlink}
\usepackage{todonotes}
\usepackage{tcolorbox}

\tcbset{
    todo/.style={
        colback=red!5!white,    % Background color
        colframe=red!75!black, % Border color
        fonttitle=\bfseries,   % Title font style
        title={Note},    % Default title
        boxrule=0.5mm,         % Border thickness
    }
}

\setuptodonotes{inline}

\def\BibTeX{{\rm B\kern-.05em{\sc i\kern-.025em b}\kern-.08em
    T\kern-.1667em\lower.7ex\hbox{E}\kern-.125emX}}
\begin{document}

\title{Leveraging Log Probabilities in Language Models to Forecast Future Events}

\author{\IEEEauthorblockN{Tommaso Soru}
\IEEEauthorblockA{\textit{Serendipity AI Ltd.} \\ \textit{London, United Kingdom} \\
tom@serendipity.ai \orcidlink{0000-0002-1276-2366}}
\and
\IEEEauthorblockN{Jim Marshall}
\IEEEauthorblockA{\textit{Serendipity AI Ltd.} \\ \textit{London, United Kingdom} \\
jim@serendipity.ai}
}

\maketitle

\begin{abstract}
In the constantly changing field of data-driven decision making, accurately predicting future events is crucial for strategic planning in various sectors.
The emergence of Large Language Models (LLMs) marks a significant advancement in this area, offering advanced tools that utilise extensive text data for prediction.
In this industry paper, we introduce a novel method for AI-driven foresight using LLMs.
Building on top of previous research, we employ data on current trends and their trajectories for generating forecasts on 15 different topics.
Subsequently, we estimate their probabilities via a multi-step approach based on log probabilities.
We show we achieve a Brier score of 0.186, meaning a +26\% improvement over random chance and a +19\% improvement over widely-available AI systems.
\end{abstract}


\begin{tcolorbox}[todo]
A shorter version of this ongoing work was accepted as a position paper at the 19th IEEE International Conference on Semantic Computing (ICSC 2025) with the title \textit{Anticipating the Future with Large Language~Models}.
\end{tcolorbox}

% [x] intro with overview
% [x] related work with short bibliography on LLM-based forecasting + AI in futures studies
% [x] 3 components + details: forecast generator, probability estimator, fact checker
% [x] the dataset
% [x] calibration using training set
% [x] evaluation of probability estimator
% [x] discussion and conclusion

\section{Introduction}

In the constantly changing field of data-driven decision making, accurately predicting future events is crucial for strategic planning in various sectors.
The emergence of Large Language Models (LLMs) marks a significant advancement in this area, offering advanced tools that utilise extensive text data for prediction.
This paper aims to examine the use of LLMs in forecast generation and probability estimation, emphasising how these models revolutionise the ability to predict future occurrences.

Over recent years, LLMs have demonstrated a remarkable ability to encode and process real-world knowledge within their vast architectures.
These models, highlighted by innovations such as Generative Pre-trained Transformers (GPT)~\cite{b3} and their derivatives, have shown exceptional proficiency in handling various natural language processing (NLP) tasks, from sentiment analysis to predictive modelling.
However, their effectiveness in event forecasting remains a contentious issue, as many experts still consider Generative AI ``hardly useful for forecasting and decision intelligence''~\cite{b10}.
We aim to challenge this view.

\begin{figure}[t]
    \centering
    \includegraphics[scale=0.24]{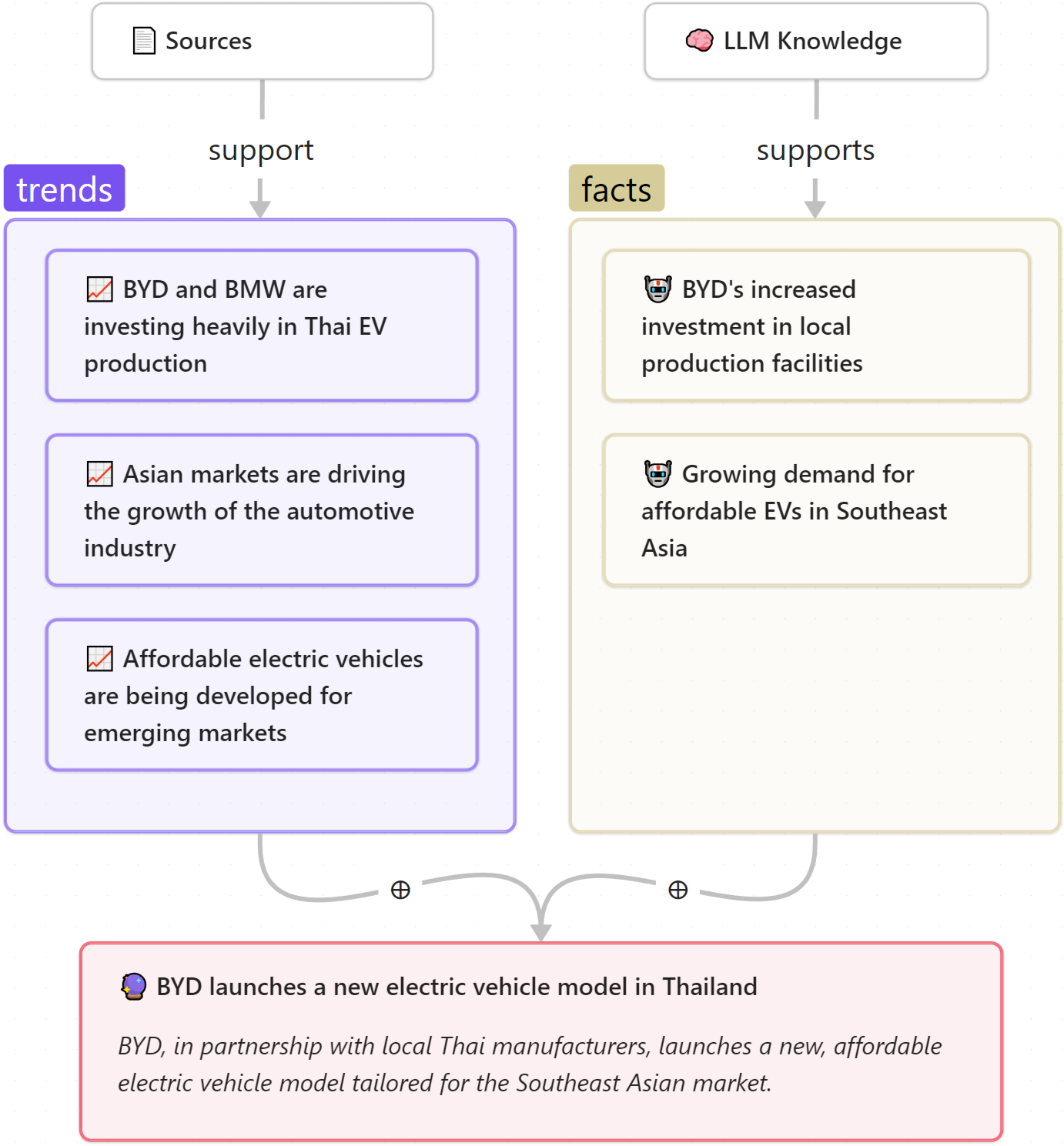}
    \caption{An example of forecast generation.}
    \label{fig:forecastgen}
\end{figure}

\begin{figure*}[t]
    \centering
    \includegraphics[scale=0.26]{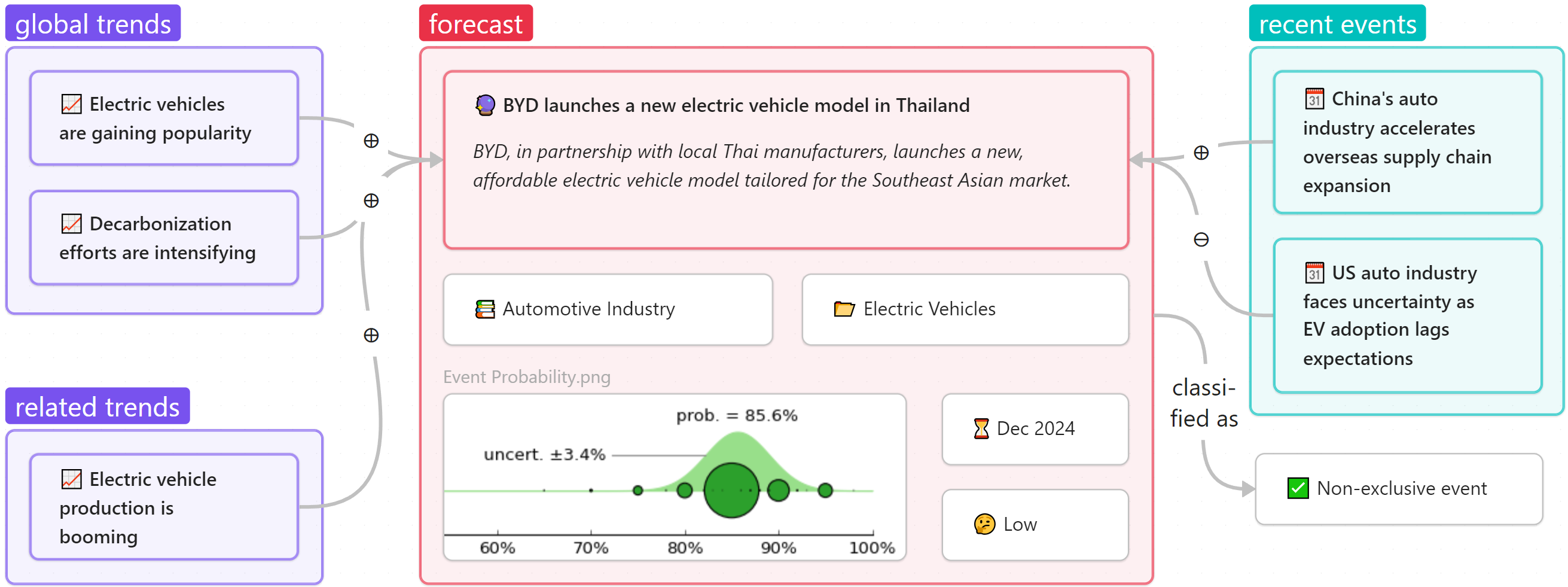}
    \caption{An example of probability estimation.}
    \label{fig:probestimation}
\end{figure*}

The scope of this paper also extends to the application of LLMs in the context of \textit{Futures Studies} -- also known as \textit{Foresight} -- a discipline that is highly focused on evaluating the plausibility of different future scenarios.
Typically, this task is performed manually by experts through detailed and comprehensive analysis of the world’s most transformative issues.
Although direct research into the use of LLMs in futures studies is limited, the potential for these models to revolutionise this field is unmistakable.
While LLMs have been used recently for event forecasting, these methods presuppose the knowledge of such events and merely estimate their probability.
Conversely, ours is the first work towards AI-powered blue-sky forecast generation based on available facts and current trends.
We backtest our generated forecasts and show we achieve -- in terms of Brier score -- a +26\% improvement over random chance and a +19\% improvement over widely-available AI systems.

This paper is organised as follows.
Related work is introduced in~\autoref{sec:related}.
% We define the terminology in~\autoref{sec:prelim}.
We describe the approach in~\autoref{sec:approach}.
We discuss results in~\autoref{sec:evaluation}.
Finally, we conclude.

\section{Related Work} \label{sec:related}

The advancement of GPT models has recently transformed natural language processing by introducing effective unsupervised learning methods that generate coherent and contextually-appropriate text~\cite{b3}.
The forecasting capability of these models on diagnosis and triage was tested using healthcare data in~\cite{b12}.
Authors in~\cite{b11} developed a retrieval-augmented language-model system designed to automatically search for relevant information about a possible future event, perform a forecast, and aggregate predictions.
Recently, \cite{b13} claimed that LLMs can achieve forecasting accuracy rivalling that of human-crowd forecasting tournaments via an ensemble method of forecast aggregation.
\textit{Reasoning and Tools for Forecasting} is a framework of reasoning-and-acting agents that can dynamically retrieve updated information and run numerical simulation with equipped tools~\cite{b14}.
A new AI forecasting system using chain-of-thought was very recently published~\cite{b15}; unfortunately, its demo is no longer available online at the time of writing this paper.

\section{Approach} \label{sec:approach}

Our system is composed of three main parts:

\begin{enumerate}
    \item a \textbf{Forecast Generator}, which generates titles, descriptions, and timeframe of potential future events from a topic;
    \item a \textbf{Probability Estimator}, which outputs a probability, an uncertainty value for a forecast, and related trends;
    \item a \textbf{Fact Checker}, used to automatically evaluate whether an event has happened within a given timeframe.
\end{enumerate}

In the following, we elaborate on and describe each component in detail.

\subsection{Forecast Generator}

The Forecast Generator is an AI-powered foresight tool whose goal is to generate potential future events and their metadata starting from as little as a topic name (e.g., \textit{Automotive}).
Building on top of previous research on LLM-driven trend extraction and analysis~\cite{b16}, we collect a broad sample of trends from recent sources.
At query time, trends are semantically filtered by relevance to the input topic.
This method updates the model's knowledge, which is typically limited to some past cut-off date.

An example of forecast generation is reported in~\autoref{fig:forecastgen}.
The model generates a forecast based on current trends and facts, which are supported by sources and LLM background knowledge, respectively.
The final forecast contains a title and a short description.

For backtesting, we created a dataset of $N=150$ forecasts from 15 different topics.
In order to ensure the absence of leakage of any future data into the dataset, all forecasts were created in February 2024 and consequently fact-checked in October 2024.
To compute its probability, each forecast is forwarded to the component introduced in the next subsection.

\subsection{Probability Estimator}

Goal of the Probability Estimator is to take a future event as input and output a probability along with a range of values.
Unlike previous systems which consider only the LLM's top completion, our system considers all possible guesses for the probability estimation.
This way, not only can we estimate a probability $\hat{P}$ as the weighted average of all guesses $P_i$, but also we can calculate the uncertainty $\hat{U}$ of a given answer as the weighted standard deviation of all guesses:

\begin{equation}
    \hat{P} = \frac{\sum_{i=1}^n e^{w_i} P_i}{\sum_{i=1}^n e^{w_i}}
\end{equation}

\begin{equation}
    \hat{U} = \sqrt{\frac{\sum_{i=1}^n e^{w_i} (P_i - \hat{P})^2}{\sum_{i=1}^n e^{w_i}}}
\end{equation}

where $w_i$ is the logarithmic probability (in short, \textit{logprob}) of the token that represents probability $P_i$.
To the best of our knowledge, ours is the first LLM-based forecasting approach that exploits logprobs to compute the final probability value.

Besides their knowledge not being up-to-date, two other limitations of LLMs are their incoherence and inconsistency across runs.
\textit{Incoherence} in forecasting happens when two forecasts made for the same event at the same time are assigned different probabilities.
To address this issue, we maximise completion determinism by setting temperature and top P to the minimum values allowed by the model.
After applying these settings, we noticed that the divergence of probability values across runs remained within 1\%.
\textit{Inconsistency} can be found when estimating probabilities for events that are mutually exclusive, e.g. elections or sports events.
Specifically, we ensure that the sum of the probabilities for all outcomes (e.g., each one of the candidates or competing teams winning) does not exceed 100\%.
Additional checks may be introduced to avoid ontological inconsistencies, such as having $P(X) < P(Y)$ with $X$ semantically entailed in $Y$.

The Probability Estimator pipeline expects six steps:

\begin{enumerate}
    \item reformulate and categorise query, so that the event is spelled correctly in English;
    \item search global and local trends relevant to the query;
    \item collect sources via semantic search over a news provider;
    \item extract the most important events from the headlines;
    \item check whether the event is non-exclusive;
    \item estimate event probability and uncertainty.
\end{enumerate}

The probability estimation for our running example can be seen in~\autoref{fig:probestimation}.
Given the event and timeframe obtained during forecast generation, the model outputs a probability and an uncertainty value, as well as the trends and recent events which positively or negatively affect its probability.
Each bubble in the smaller chart represents an estimate of the probability, with its size proportional to the logprob of that estimate.
% We also differentiate between \textit{exclusive} and \textit{non-exclusive} events.

\subsection{Fact Checker}

The Fact Checker component takes a potential event and a timeframe as input.
It outputs three possible values: \textit{Happened} ($1$) if the event has happened, \textit{Inconclusive} ($0$) if there is no sufficient evidence to confirm the event, or \textit{Did not happen} ($-1$) if there are reasonable grounds to say that the event cannot possibly have happened.
As in the Probability Estimator, relevant sources are collected from our news provider and forwarded to the model to fact-check the input event.
We manually evaluated the reliability of the Fact Checker; having achieved a perfect score of $150/150$ on our dataset, it will be employed to track future forecasts automatically.

\subsection{Model calibration}

To use the knowledge of forecast uncertainties to our advantage, we learn a transformation function between LLM-returned values and the final probability output.
After discarding invalid forecasts, i.e. those events which the Fact Checker found they had already happened at the time of forecasting, we split the dataset equally into training and test set, retaining a test set of $N=72$ elements.
We employ a \textit{Support Vector Regression} model to find the optimal transformation function and ultimately calibrate the model.

\begin{figure}[t]
    \centering
    \includegraphics[scale=0.55]{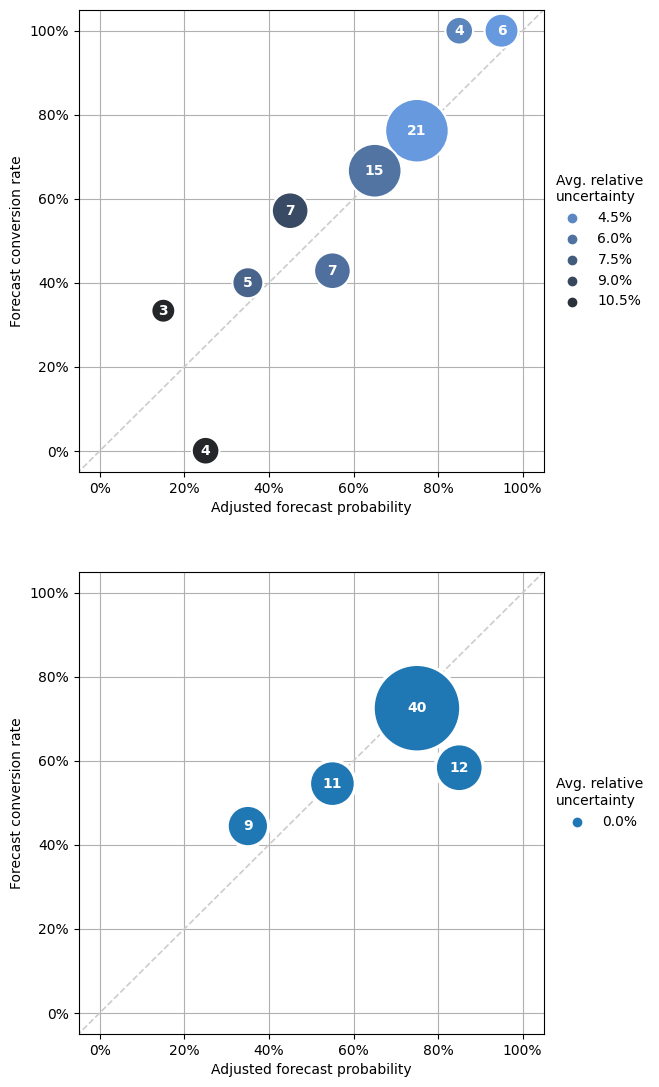}
    \caption{Calibration plot for our Probability Estimator (above) and vanilla GPT-4o (below).}
    \label{fig:calibs}
\end{figure}

% \begin{figure}[h]
%     \centering
%     \includegraphics[scale=0.55]{calibration-plot-gpt.png}
%     \caption{Calibration plot for vanilla GPT-4o.}
%     \label{fig:calibgpt}
% \end{figure}

\section{Evaluation} \label{sec:evaluation}

The trends were extracted and analysed from a collection of $\sim$45,000 articles over 6 months ending February 2024.
As previously mentioned, the time between forecasting and fact-checking is about 240 days.
We could not find any system to compare against, therefore we use a widely-available AI as baseline.
We employ OpenAI's \textit{GPT-4o} model~\cite{b17} in both our Probability Estimator and the baseline.
As in previous literature~\cite{b11,b13,b14}, we use Brier scores to quantify the performance of our Probability Estimator:

\begin{equation}
    B = \frac{1}{N} \sum_{i=1}^N {(f_i - o_i)^2}
\end{equation}

where $N$ is the test set size, $f_i$ is the forecasted probability of each event $i$, and $o_i \in \lbrace 0, 1\rbrace$ its outcome.

Calibration plots can be seen in \autoref{fig:calibs}.
Each bubble represents a subset of forecasts, grouped by estimated probability; the number on each bubble is the size of the subset.
Our plot reports shades of colour, where light means low uncertainty and dark means high uncertainty.
The correlation between the probability returned by the models and the forecast conversion rate, i.e. the actual rate of forecasts turning true, can be seen in both charts, although it is higher in ours.

\begin{table}[h]
\centering
\caption{Evaluation results for the two approaches. Lower is better. Other Brier scores are given for reference.}
\label{tab:results}
\begin{tabular}{lc}
\toprule
\textbf{Reference}           & \textbf{Brier score} \\
\midrule
Humans                       & 0.200–0.300 \\
Prediction markets           & 0.100–0.200 \\
Manifold (480 days)          & 0.182       \\
Average superforecasters     & 0.150       \\
\midrule
\textbf{Evaluation results}  & \textbf{Brier score} \\
\midrule
Random chance                & 0.250       \\
Vanilla GPT-4o               & 0.236       \\
Probability Estimator (ours) & 0.186       \\
\bottomrule
\end{tabular}
\end{table}

\autoref{tab:results} shows the evaluation results for our Probability Estimator against vanilla GPT-4o.
We achieve a Brier score of 0.186, which means a +19\% improvement over the widely-available AI and a +26\% improvement over random chance.
These results confirm that the knowledge condensed in the trends' trajectories as well as the ability of estimating uncertainties do provide strategic advantage.

The table also features additional reference ranges from Calibration City\footnote{\url{https://calibration.city/}}.
According to the evaluation, our Probability Estimator could score better than some prediction markets and would be only 0.036 Brier points away from the average superforecaster.
Moreover, Calibration City reports forecasts whose market open length is between 0 and 480 days and uses the probability at market midpoint as the value to test.
As the time between forecasting and fact-checking in our dataset is about 240 days, we would need to compare with markets twice as long.
The only platform having an analogous magnitude of data ($N=39$) for 480-day-long markets is Manifold.
With a Brier score of $0.182$, this result is in line with the score obtained by our Probability Estimator on our own dataset.
Although these last results are not directly comparable, we believe they are significant and strongly argue in favour of autonomous LLM-based forecasting.

\begin{figure}[t]
    \centering
    \includegraphics[scale=0.55]{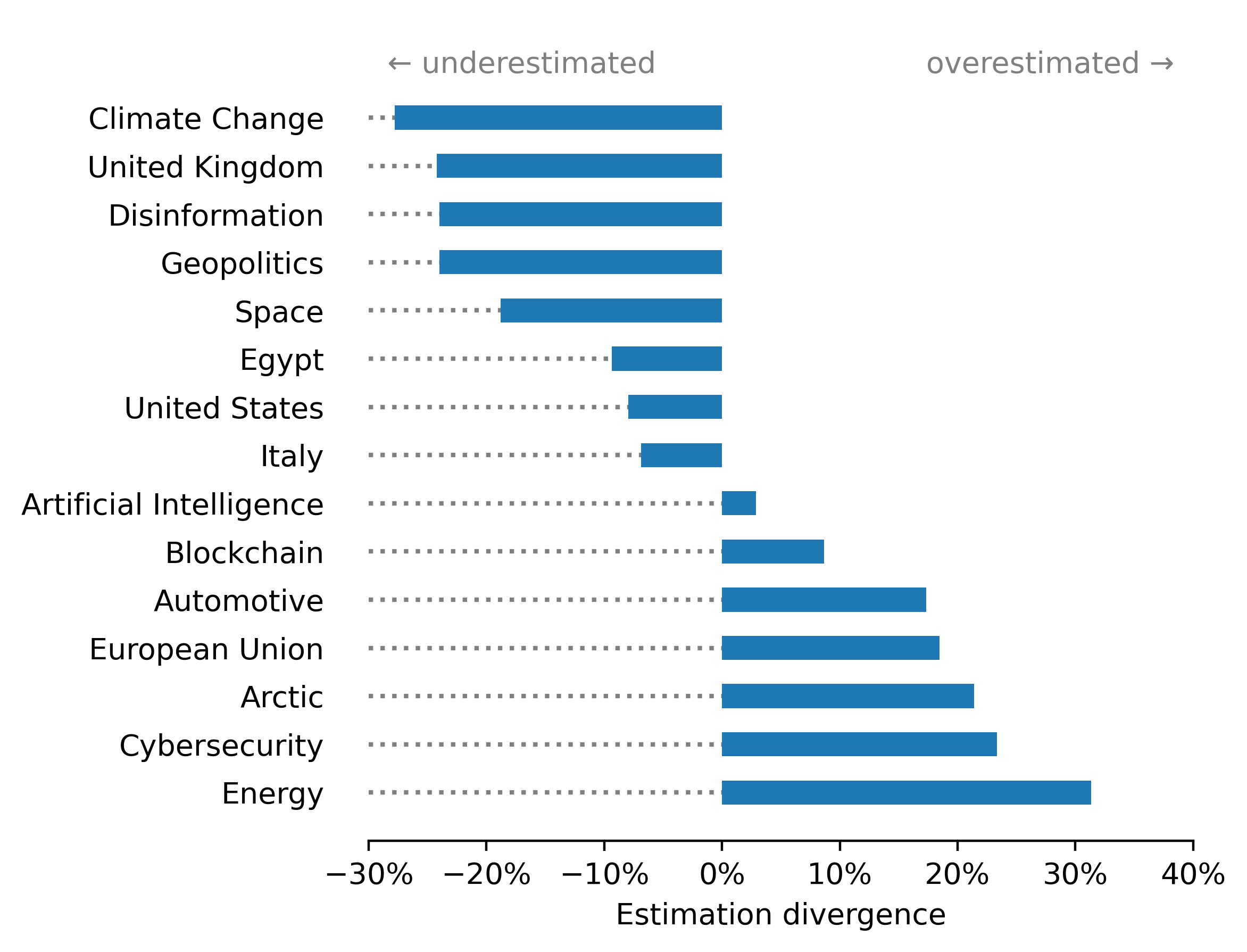}
    \caption{Probability estimation performance by topic.}
    \label{fig:topics}
\end{figure}

In the future, even the semantic dimension may be used to calibrate probabilities better.
\autoref{fig:topics} shows that for topics such as \textit{Climate Change}, forecast probabilities were underestimated, as opposed to those in topics such as \textit{Energy} which were overestimated.

\section{Conclusion} \label{sec:conclusion}

In this paper, we have introduced a novel approach to AI-driven foresight based on current trends, LLMs, and their logarithmic probabilities.
Despite recent controversies regarding the ability of language models to perform forecasts accurately, we believe there is now enough evidence for this possibility.
We found our systems are not yet performing at the same level as superforecasters.
However, our results confirm that the knowledge condensed in trends and uncertainty values do provide strategic advantage to the end user.

This work may enable several applications in strategic foresight and adjacent fields.
After testing the quality of the returned probabilities, future evaluations may target the utility of the forecast themselves.
By exploiting the visionary capability of our Forecast Generator, we can create complex multi-forecast scenarios and analyse their interactions with the environment.
Simulations by LLM-powered agents can become more reliable having a system which can compute probabilities of virtually any kind of event.
Finally, new avenues may be opened for data-driven decision making in a personal, enterprise, or public setting.

\end{document}